\documentclass[conference]{IEEEtran}

\usepackage{graphics} 
\usepackage{epsfig} 
\usepackage{amsmath} 
\usepackage{amssymb}  
\usepackage{etoolbox}

\def\BibTeX{{\rm B\kern-.05em{\sc i\kern-.025em b}\kern-.08em
    T\kern-.1667em\lower.7ex\hbox{E}\kern-.125emX}}

\begin{document}

\title{Transfer learning optimization based on evolutionary selective fine tuning}

\author{Jacinto Colan$^{1}$, Ana Davila$^{2}$ and Yasuhisa Hasegawa$^{2}$
\thanks{$^{1}$\quad
Department of Micro-Nano Mechanical Science and Engineering, Nagoya University, Furo-cho, Chikusa-ku, Nagoya, Aichi 464-8603, Japan}
\thanks{$^{2}$\quad
Institutes of Innovation for Future Society, Nagoya University, Furo-cho, Chikusa-ku, Nagoya, Aichi 464-8601, Japan}
\thanks{Correspondence: {\tt\small colan@robo.mein.nagoya-u.ac.jp}}
\thanks{
This work was supported in part by the Japan Science and Technology Agency (JST) CREST under Grant JPMJCR20D5, and in part by the Japan Society for the Promotion of Science (JSPS) Grants-in-Aid for Scientific Research (KAKENHI) under Grant 22K14221.}
}


\maketitle

\begin{abstract}
Deep learning has shown substantial progress in image analysis. However, the computational demands of large, fully trained models remain a consideration. Transfer learning offers a strategy for adapting pre-trained models to new tasks. Traditional fine-tuning often involves updating all model parameters, which can potentially lead to overfitting and higher computational costs. This paper introduces BioTune, an evolutionary adaptive fine-tuning technique that selectively fine-tunes layers to enhance transfer learning efficiency. BioTune employs an evolutionary algorithm to identify a focused set of layers for fine-tuning, aiming to optimize model performance on a given target task. Evaluation across nine image classification datasets from various domains indicates that BioTune achieves competitive or improved accuracy and efficiency compared to existing fine-tuning methods such as AutoRGN and LoRA. By concentrating the fine-tuning process on a subset of relevant layers, BioTune reduces the number of trainable parameters, potentially leading to decreased computational cost and facilitating more efficient transfer learning across diverse data characteristics and distributions.
\end{abstract}


\section{INTRODUCTION}

Despite the success of deep learning in various image analysis applications, the requirement for large annotated datasets remains a significant challenge in many specialized fields that require expert annotation or involve sensitive information, such as medical imaging. This data scarcity motivates the development of techniques that can leverage existing labeled data more effectively.

Transfer learning offers a practical solution to the limited data challenge by adapting pre-trained models from data-rich domains to specific target tasks. This approach reduces the need for extensive labeled datasets while improving training efficiency and classification performance. However, effective transfer learning faces substantial challenges when source and target domains differ significantly, potentially leading to negative transfer or catastrophic forgetting, where the adapted model performs worse than baseline training.
Fine-tuning, a common transfer learning approach, involves adjusting pre-trained model parameters using task-specific data. Although this method exploits useful feature representations, it presents several technical challenges, including the potential deterioration of pre-trained features and decreased performance on out-of-distribution data, especially when domains vastly differ \cite{kumar22fine}. A common practice in fine-tuning is the selective adjustment of layers. However, selecting which layers to fine-tune adds complexity. Traditionally, efforts have focused on training only the terminal layers, but recent studies suggest that adaptive layer selection strategies might be more effective \cite{lee23surgical}. Furthermore, additional fine-tuning parameters such as learning rates and weight decay significantly impact model performance and stability \cite{alshalali18fine}, often requiring domain-specific expertise \cite{nguyen21fine}.

To address these challenges, the fine-tuning process can be framed as an optimization problem, thereby enabling the application of nature-inspired algorithms. These algorithms have demonstrated efficacy in tackling complex optimization tasks, including constrained inverse kinematics \cite{davila24realtime}, hyperparameter tuning \cite{vincent23improved}, and feature selection \cite{nssibi23advances}. Among these, Evolutionary Algorithms (EA) and Swarm Intelligence (SI) approaches each have their own distinct strengths. EAs outperform in maintaining population diversity, enhancing solution robustness, while SI algorithms are known for their efficiency and simplicity \cite{vesterstrom04comparative}. Hybrid algorithms that combine elements from both EAs and SI aim to leverage their collective strengths, balancing the exploration of new solutions with the exploitation of known good configurations \cite{grosan07hybrid}.

In this paper, we introduce an adaptive fine-tuning method using an improved evolutionary algorithm to explore fine-tuning configurations. The proposed approach automatically determines layer freezing strategies and optimizes learning rates for active layers, achieving efficient exploration through systematic evaluation of target dataset subsets during each evolutionary generation.
The main contributions of this paper are as follows.

\begin{itemize}
    \item  A novel evolutionary-based adaptive fine-tuning approach that dynamically optimizes learning rates and layer freezing strategies across diverse datasets and architectures.
    \item Comprehensive comparison against existing fine-tuning methods across nine image classification datasets and four CNN architectures, demonstrating improved performance and adaptability.
\end{itemize}

\section{Proposed method}
\label{sec:2}

\begin{figure}[hbt]	
\centering
\includegraphics[width=\linewidth]{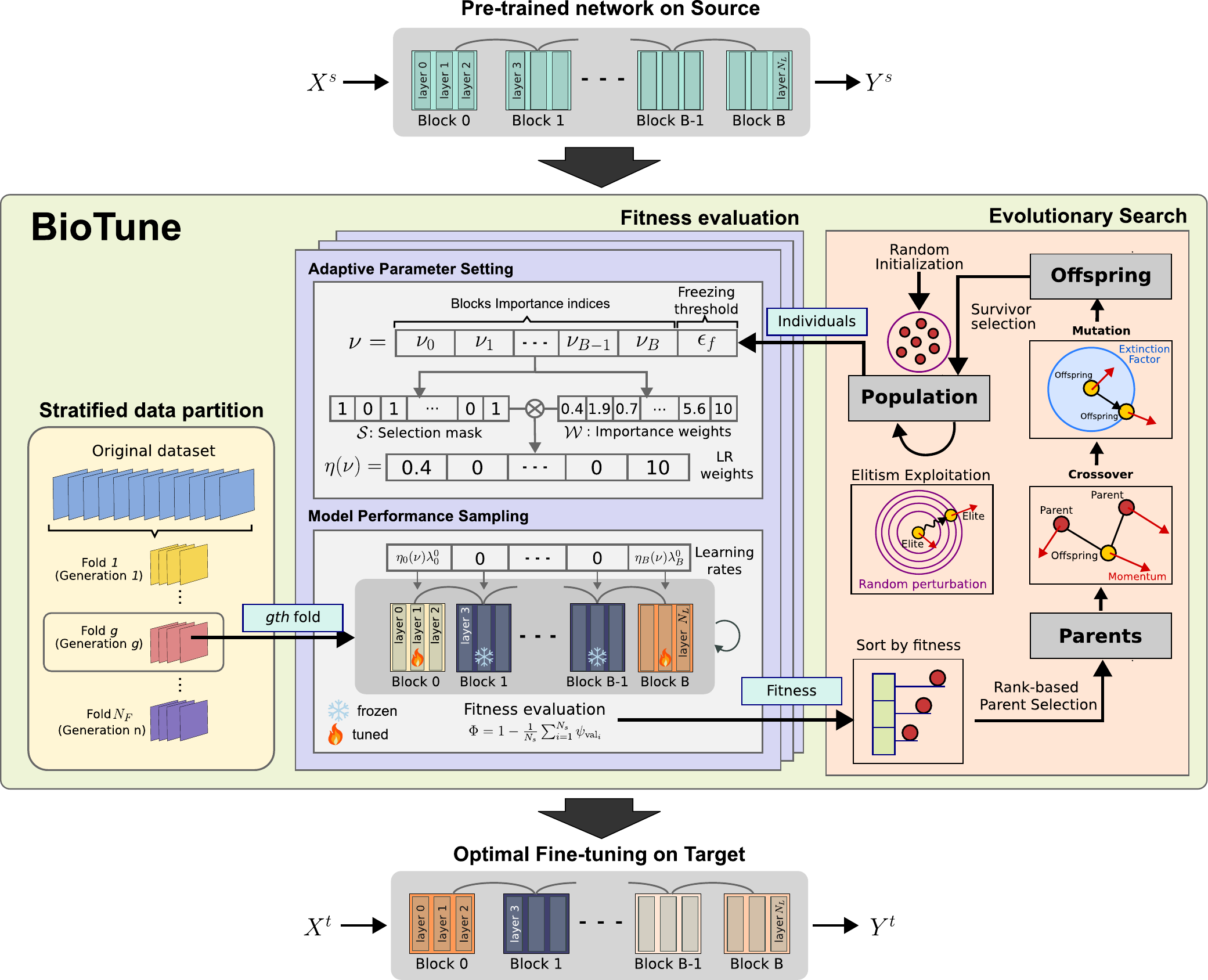}
\caption{Overview of BioTune \label{fig:1}}
\end{figure}

We propose a novel fine-tuning framework for selective and adaptive fine-tuning, which simultaneously optimizes the set of layers to fine-tune and their corresponding learning rates to maximize classification performance. Our approach leverages an improved evolutionary optimization strategy to guide the exploration process \cite{starke19memetic}. An overview of the proposed method is presented in Figure \ref{fig:1}. The BioTune method comprises the following stages: model selection and pre-training, stratified data partitioning, evolutionary search, fitness evaluation, and fine-tuning with optimal configurations.

\subsection{Fine-tuning optimization problem}
\label{sec:3.1}
Given a pre-trained model $M = \{m_b: b \in \{0, \dots, B\}\} $, composed of $B+1$ sets of layers or blocks grouped based on their function (e.g., layers belonging to a residual block or a fully connected classifier), pre-trained on a source dataset with abundant labeled images $\mathcal{X}_s = \{(x_i^s, y_i^s)\}_{i=1}^{N_s} $, our goal is to determine an optimal fine-tuning configuration $\nu^* $ that maximizes the accuracy of the network on a target dataset $\mathcal{X}_t = \{(x_i^t, y_i^t)\}_{i=1}^{N_t} $, which contains different classes. The optimization problem can be framed as:

\begin{equation}
    \nu^* = \underset{\nu}{\arg\max} \, \text{Acc}(M(\omega^0), \lambda(\nu), \mathcal{X}_t)
\end{equation}

where $\omega^0$ represents the model pre-trained parameters, and $\lambda(\nu)$ defines the learning rates for each block during fine-tuning, given by:

\begin{equation}
\label{eq:2}
    \lambda(\nu) = \left\{ \eta_b(\nu) \lambda^0_b : b \in \{0, \dots, B\} \right\}
\end{equation}

where $\lambda^0_b $ is the predefined base learning rate for each block $b$, and $\eta_b(\nu) $ is a weight that adjusts the base learning rate according to the importance of each block during fine-tuning.
The function $\eta_b(\nu)$ controls fine-tuning intensity by enabling block-level selection and learning rate assignment. For crucial adaptation layers, $\eta_b(\nu) > 1$ increases learning rates for significant updates; for less important layers, $\eta_b(\nu) < 1$ reduces rates for minimal adjustments; and for non-contributing layers, $\eta_b(\nu) = 0$ freezes parameters, reducing computational costs and accelerating fine-tuning. The optimal configuration $\nu^*$ is determined through evolutionary search that iteratively explores the configuration space.

\begin{table*}[ht]
    \newrobustcmd{\B}{\bfseries} 
    \newrobustcmd{\U}{\underline}
    \centering
    \renewcommand{\arraystretch}{1.4} 
    \caption{Comparison of accuracy with standard errors across three runs for various fine-tuning methods and datasets. For BioTune, the second line shows the relative performance improvement compared to FT. Top score are highlighted in bold.}
    \scalebox{0.83}{
    \begin{tabular}{r | c c c | c c | c c | c c}
        \cline{1-10}
                    & \multicolumn{3}{c|}{Digits}  	& \multicolumn{2}{c|}{Objects} & \multicolumn{2}{c|}{Fine-grained} & \multicolumn{2}{c} {Specialized}    \\ 
        Method & MNIST         & USPS          &SVHN           & CIFAR-10  & STL-10       & Flowers-102    & FGVC-Aircraft  & DTD       & ISIC2020   \\ \hline
        FT         & 98.96 (0.0)  & 97.05 (0.1)  & 95.56 (0.2)  & 95.65 (0.1)  & 97.33 (0.0)  & 85.33 (0.5)  & 58.68 (1.9)  & 68.03 (0.1)  & 78.91 (0.7)  \\
        \cline{1-10}
        LP \cite{kumar22fine}           & 92.53 (0.2)  & 91.78 (0.0)  & 44.08 (0.1)  & 81.03 (0.0)  & 96.86 (0.1)  & 82.72 (0.6)  & 33.61 (0.4)  & 66.00 (0.1)  & 77.49 (0.5)  \\
        $L^1$-$SP$ \cite{li18explicit}  & 98.98 (0.1)  & 97.31 (0.0)  & 96.01 (0.0)  & 95.60 (0.0)  & 97.01 (0.0)  & 87.82 (0.5)  & 60.55 (1.9)  & 68.52 (0.1)  & 80.62 (1.2)  \\
        $L^2$-$SP$ \cite{li18explicit}  & 98.87 (0.0)  & 97.00 (0.1)  & 95.47 (0.1)  & 95.78 (0.1)  & 97.20 (0.1)  & 85.29 (0.5)  & 61.56 (1.1)  & 69.01 (0.2)  & 79.77 (2.0)  \\
        G-LF \cite{howard18universal}        & 98.82 (0.1)  & 96.72 (0.2)  & 94.00 (0.0)  & 93.77 (0.0)  & 97.32 (0.0)  & 87.59 (0.9)  & 54.77 (1.3)  & 67.85 (0.3)  & 77.49 (1.2)  \\
        G-FL \cite{mukherjee20distilling}        & 98.57 (0.0)  & 96.86 (0.1)  & 94.91 (0.0)  & 95.43 (0.0)  & 97.01 (0.0)  & 86.14 (0.2)  & 49.22 (0.7)  & 65.42 (0.3)  & 76.92 (1.3)  \\
        AutoRGN \cite{lee23surgical}     & 99.00 (0.0)  & 96.91 (0.2)  & \B{96.08} (0.0)  & 96.05 (0.0)  & 96.92 (0.1)  & 85.5 (0.3)  & 57.94 (0.8)  & 65.70 (0.2)  & 79.48 (0.4)  \\
        LoRA \cite{hu21lora}        & 98.51 (0.1)  & 96.92 (0.0)  & 95.46 (0.1)  & 95.17 (0.1)  & 97.46 (0.1)  & 86.01 (0.2)  & 54.78 (1.3)  & 68.17 (0.4)  & 80.91 (1.0)  \\
        \cline{1-10}
        BioTune & \B{99.13} (0.0) & \B{97.57} (0.1) & 95.85 (0.0) & \B{96.09} (0.1) & \B{97.50} (0.0) & \B{91.68} (0.1) & \B{64.40} (0.6) & \B{69.27} (0.6) & \B{82.90} (0.8) \\[-1.5ex]
        (Ours) & \footnotesize{+0.2\%} & \footnotesize{+0.5\%} & \footnotesize{+0.3\%} & \footnotesize{+0.5\%} & \footnotesize{+0.2\%} & \footnotesize{+6.7\%} & \footnotesize{+9.7\%} & \footnotesize{+1.8\%} & \footnotesize{+5.1\%} \\[1pt]
        \hline
    \end{tabular}}
    \label{tab:1}
\end{table*}

\subsection{Evolutionary search}
\label{sec:3.2}

Fine-tuning configurations $\nu $ are encoded as genotypes, where the genotype of the $s $-th individual is represented as $\nu^s = \left[\nu^s_0, \nu^s_1, \dots, \nu^s_{B}, \nu^s_{B+1} \right]$. Each gene $\nu^s_b$, where $b \in {0, \dots, B}$, corresponds to an importance index for block $b$ and ranges from 0 to 1. The final gene, $\nu^s_{B+1} = \epsilon_f^s$, acts as a freezing threshold, differentiating blocks to be frozen from those to be fine-tuned. This threshold, $\epsilon_f^s$, is an optimization parameter that promotes diversity within the population \cite{dusan18novel}.

\begin{equation}
    \nu_b^{0,\dots,N_p-1} = U_{[0,1]} \quad \forall b = 0,\dots,B+1
\end{equation}

Each individual's fitness is evaluated using a function $\Phi$, which measures its predictive performance on a validation set. This evaluation comprises two stages: adaptive parameter setting and model performance sampling. 

\subsubsection{Adaptive parameter setting}

For a given individual $\nu $ being evaluated, the learning rate weights $\eta(\nu_b)= \mathcal{S} \cdot \mathcal{W} $ for each block $b $ are determined by the product of a selection mask $\mathcal{S} $ and an importance weight $\mathcal{W}$ vectors:

The selection mask $\mathcal{S} $ controls whether the corresponding set of layers is enabled for fine-tuning or remains frozen, based on the gene value $\nu_b $ and the freezing threshold $\epsilon_f$. Specifically, the mask is defined as:

\begin{equation}
    \mathcal{S}_b = \begin{cases} 
    0, & \text{if } \nu_b \leq \epsilon_f \\
    1, & \text{if } \nu_b > \epsilon_f 
    \end{cases}
\end{equation}

Here, $\mathcal{S}_b = 0 $ indicates that the layers in block $b $ are frozen, while $\mathcal{S}_b = 1 $ enables fine-tuning for that block. The threshold $\epsilon_f$ serves as the cutoff value to determine whether or not a block contributes to the fine-tuning process.
The importance weights $\mathcal{W} $ are calculated as follows:

\begin{equation}
    \mathcal{W}_b = 10^{2(\nu_b - 0.5)}
\end{equation}

This formulation allows $\mathcal{W} $ to exponentially scale the learning rates within [0.1,10], emphasizing layers with higher importance while de-emphasizing those with lower importance.
Finally, the learning rate $\lambda_b $ for each block $b $ is computed using Eq.~\ref{eq:2}.

\subsubsection{Model performance sampling}

For each individual, the corresponding learning rate configuration is applied to the pre-trained model.  If $\eta(\nu_b)$ for a block is 0, its parameters are frozen. Otherwise, the block's learning rate is scaled by its corresponding weight $\eta(\nu_b)$, enabling differential parameter updates during training via gradient-based optimization.The model is then fine-tuned on the target training set for a fixed number of epochs using categorical cross-entropy loss. To ensure robustness, the process is repeated with $N_s$ different random seeds. The average of the validation accuracies from these trials is used as the fitness metric $\Phi(\nu) = 1 - \frac{1}{N_s} \sum_{i=1}^{N_s} \psi_{\text{val}_i}(\nu)$. Here, $\psi_{{val}_i} $ is the accuracy obtained in the validation set for the $i $-th trial. The population is then sorted according to their fitness values, and the top performers are selected as parents for the next generation. Evolutionary operations are then applied to these parents to generate offspring: exploitation, crossover, mutation and adaptation. 

The offspring are subsequently evaluated, and those with the highest fitness are selected to form the new population. This cycle repeats over multiple generations until the optimal fine-tuning configuration is identified, as indicated by the highest prediction accuracy on the validation set. This optimal configuration is then used to fine-tune the model with the full training set, which is subsequently evaluated on the held-out test set of the target dataset.

\subsection{Stratified Data Partitioning}
\label{sec:3.3}

To address the computational demands of evolutionary search, we employ stratified data partitioning. The training dataset is divided into $N_s$ stratified folds $f_g = g \bmod N_s$. Each fold preserves the original class distribution, and fold $f_g$ is used for evaluation at generation $g$. This reduces per-generation computational cost and ensures robust exploration by exposing candidates to different training samples across generations. 

\begin{table*}[ht]
    \newrobustcmd{\B}{\bfseries} 
    \newrobustcmd{\U}{\underline}
    \centering
    \renewcommand{\arraystretch}{1.4} 
    \caption{Percentage of trainable parameters in ResNet-50 for each dataset}
    \scalebox{0.83}{
    \begin{tabular}{r | c c c | c c | c c c c}
        \cline{1-10}
                    & \multicolumn{3}{c|}{Digits}  	& \multicolumn{2}{c|}{Objects} & \multicolumn{4}{c}{Specialized}    \\ 
        Method & MNIST         & USPS          &SVHN           & CIFAR-10  & STL-10       & Flower-102    & FGVC-Aircraft  & DTD       & ISIC2020   \\ \hline
        BioTune  & 29.97  & 36.86  & 100.0   & 100.0   & 64.93   & 99.12   & 99.96   & 64.89   & 29.93   \\
        \hline
    \end{tabular}}
    \label{tab:2}
\end{table*}

\section{Experimental validation}
\label{sec:3}

\subsection{Datasets}
\label{sec4:1}

To comprehensively evaluate the performance and generalization capabilities of our approach, we utilized a diverse set of image classification datasets: Flowers-102 \cite{nilsback08automated}, MNIST \cite{lecun98gradient}, USPS \cite{hull94database}, SVHN \cite{netzer11reading}, CIFAR-10\cite{krizhevsky09learning}, STL-10  \cite{coates11analysis}, FGVC-Aircraft \cite{maji13fine}, DTD \cite{cimpoi14describing}, ISIC2016 \cite{rotemberg21patient}. These datasets span a wide range of domains, including classification on digits, natural objects, fine-grained, and specialized domains.

\subsection{Experimental setup and implementation}
\label{sec4:2}
To evaluate our approach, we used a ResNet-50 architecture  \cite{he16deep} pre-trained on ImageNet, implemented in PyTorch 2.0 on an NVIDIA RTX A6000 GPU. For BioTune, we used a population of 10 with 3 elite individuals, 10 maximum generations, and 3 random seeds per fitness evaluation. BioTune training was limited to 30 epochs with 3 epochs patience for early stopping, and a 0.25 random perturbation for exploration. We deliberately excluded data augmentation to focus solely on the fine-tuning impact, and implemented various established fine-tuning techniques for comprehensive comparison \cite{davila24comparison}.

\subsection{Performance on diverse image classification datasets}
\label{sec5:1}

The experimental results of the proposed BioTune model across the diverse set of image classification tasks are presented in Table~\ref{tab:1}, where we utilize test set accuracy as the primary performance metric. We report the mean accuracy across three independent runs along with standard errors (shown in parentheses). For BioTune specifically, we evaluate the top-5 performing configurations identified across all explored generations and report the highest achieved accuracy.

The results show that BioTune consistently achieves competitive or superior performance compared to existing fine-tuning methods across most datasets, with SVHN being the only exception. BioTune demonstrates moderate improvements (around 0.5\%) over the FT baseline for datasets closely aligned with the source domain of natural images. More substantial gains are observed on fine-grained classifications (with FGVC-Aircraft achieving a 9.7\% improvement) and specialized datasets far from the source domain, such as DTD (textures) and ISIC2020 (dermoscopy), which show improvements of 1.8\% and 5.1\%, respectively.

\begin{figure}[bt]	
\centering
\includegraphics[width=0.9\linewidth]{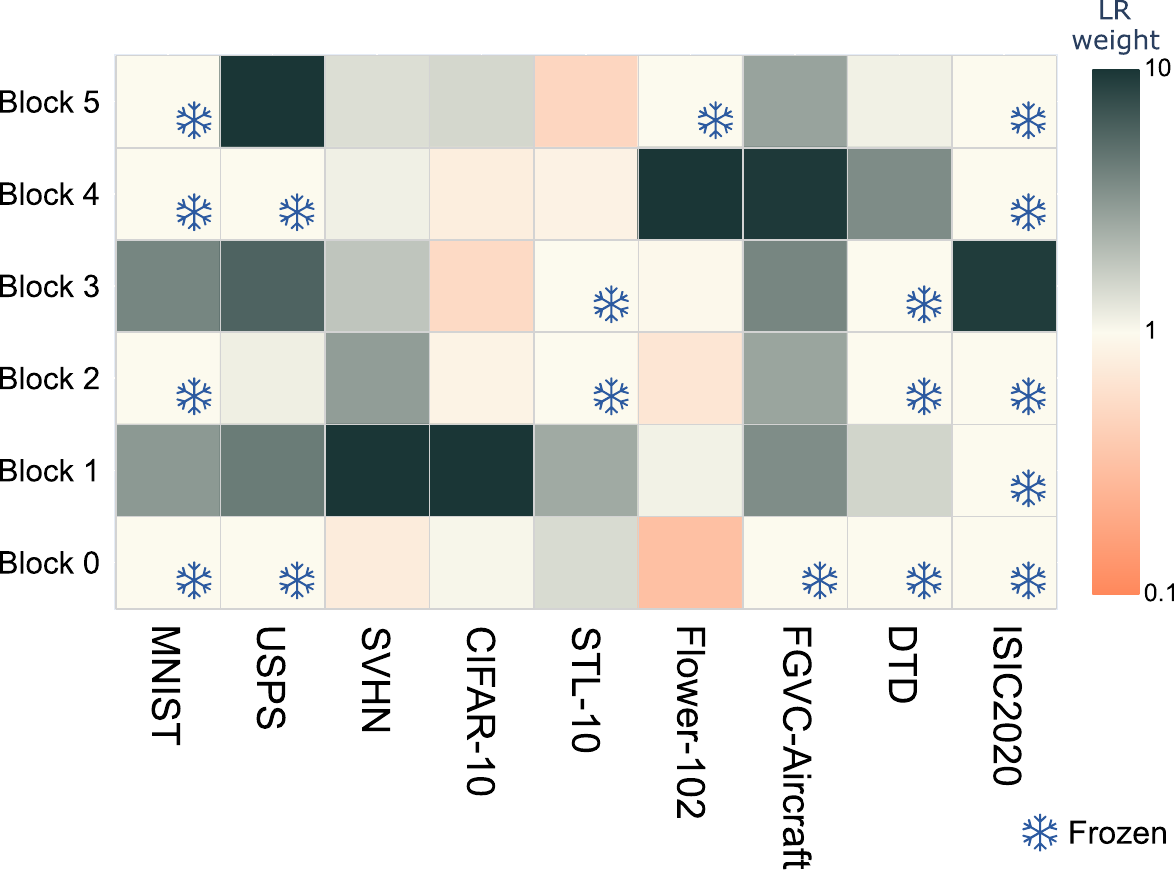}
\caption{Optimal fine-tuning configurations discovered by BioTune across datasets. Frozen layers are marked with a snowflake symbol, and learning rate weights are shown as a heatmap (0.1 to 10). \label{fig:2}}
\end{figure}  

\begin{table}[bt]
\centering
\renewcommand{\arraystretch}{1.2} 
\caption{Mean test accuracy and computation time for different training data percentages}
\scalebox{1.0}{
\begin{tabular}{l  c c  c c c}
\cline{1-6}
                    & \multicolumn{5}{c}{Percentage of Training dataset}  \\ 
Dataset        
& 10\%          & 20\%          & 25\%          & 50\%          & 100\%    \\ \hline
Mean Accuracy       & 90.46   & 90.34     & 90.53    & 90.95    & 91.1 \\ 
Comp. time (hours)    & 1.6   & 2.2     & 3.7    & 6.0    & 11.4 \\ 
\hline
\end{tabular}}
\label{tab:3}
\end{table}

Figure~\ref{fig:2} provides a comprehensive visualization of BioTune's optimal fine-tuning configurations for each dataset, with frozen layers denoted by snowflake symbols and learning rate weights shown as a heatmap (values 0.1-10), revealing distinct layer-wise adaptation patterns across different transfer learning scenarios. Table~\ref{tab:2} complements this by presenting the percentage of trainable parameters utilized by BioTune relative to the total available parameters in ResNet-50, showing significant variations in parameter utilization across tasks and demonstrating BioTune's ability to automatically determine the optimal parameter subset for fine-tuning.

We explored how varying the percentage of training data per generation affects BioTune's performance, highlighting the trade-off between computational cost and performance gains. Table~\ref{tab:3} summarizes these findings for the Flowers-102 dataset, demonstrating that high test accuracy ($>$90\%) is achievable using only 10\% of training data with minimal computation time (1.6 hours). While increasing training data improves performance and reduces variance, it significantly raises computational costs—with peak accuracy of 91.1\% using 100\% of training data requiring 11.4 hours. For our experimental validation, we selected a balanced 50\% split to optimize the accuracy-efficiency trade-off.

\section{CONCLUSIONS}

We introduced BioTune, a novel evolutionary-based adaptive fine-tuning approach that optimizes transfer learning by dynamically determining layer freezing and learning rates. Our comprehensive benchmarking across diverse datasets demonstrated BioTune's superior performance compared to existing methods, achieving higher accuracy while improving efficiency through selective layer freezing and reduced trainable parameters. BioTune offers a useful tool for developing more robust deep learning models, enabling more effective transfer learning across a wider range of applications and domains.

\addtolength{\textheight}{-11cm}   




\bibliographystyle{IEEEtran}
\bibliography{biblio}

\vspace{12pt}

\end{document}